\documentstyle[formulas,colacl]{article}

\input{psfig}

\title{An Empirical Investigation of Proposals in Collaborative Dialogues} 
\author{
Barbara Di Eugenio \hspace*{1cm} Pamela W.\ Jordan\\ 
\hspace*{1.3cm} {\bf Johanna D. Moore}
\hspace*{1.35cm} {\bf Richmond H.\ Thomason} \\[0.05cm] Learning
Research \& Development Center, and Intelligent Systems Program\\
University of Pittsburgh\\ Pittsburgh, PA 15260, USA \\ {\tt
\{dieugeni,jordan,jmoore,thomason\}@isp.pitt.edu} } 
\date{}

\begin{document}

\maketitle
\bibliographystyle{/users/intgen/tex/acl}

\begin{abstract} We describe a corpus-based
investigation of proposals in dialogue.
First, we describe our DRI compliant coding scheme and
report our inter-coder reliability results. Next, we test several
hypotheses about what constitutes a well-formed proposal.
\end{abstract}

\section{Introduction}\label{intro}


Our project's long-range goal (see
http://www.isp.\linebreak[4]pitt.edu/\~{ }intgen/)
is to create a unified architecture for collaborative discourse,
accommodating both interpretation and generation.  Our computational
approach \cite{thomason-hobbs:1997a} uses a form of weighted abduction
as the reasoning mechanism \cite{hobbs-etal:1993a} and modal operators to
model context.
In this paper, we describe the corpus
study portion of our project, which is an integral part of our investigation
into recognizing how conversational participants coordinate agreement. From
our first annotation trials, we found that the recognition of
``classical'' speech acts \cite{austin:1962b,searle75-short} by coders is fairly
reliable, while recognizing contextual relationships (e.g., whether an
utterance accepts a proposal) is not as reliable. Thus,
we explore other features that can help us recognize how participants
coordinate agreement.

Our corpus study also provides a preliminary assessment of the
Discourse Resource Initiative (DRI) tagging scheme.  The DRI is an
international ``grassroots'' effort that seeks to share corpora that
have been tagged with the core features of interest to the discourse
community.  In order to use the core scheme, it is anticipated that
each group will need to refine it for their particular purposes.
A usable draft core 
scheme is now available for experimentation
(see http://www.georgetown.edu/luperfoy/Discourse-Treebank/dri-home.html).
Whereas several groups are working with the unadapted core DRI scheme
\cite{core-fall97,poesio-fall97}, we have attempted to adapt it
to our corpus and particular research questions.  


First we describe our corpus, and the issue of tracking
agreement. Next we describe our coding scheme and our intercoder
reliability outcomes.  Last we report our findings on
tracking agreement.


\section{Tracking Agreement}

\label{proposal}

Our corpus consists of 24 computer-mediated dialogues\footnote{Participants
work in separate rooms and communicate via the computer interface.
The interface prevents interruptions.}  in which two participants
collaborate on a simple task of buying furniture for the living and
dining rooms of a house (a variant of the task in \cite{walker-phd}).
The participants' main goal is to negotiate purchases; the items of
highest priority are a sofa for the living room and a table and four
chairs for the dining room.  The problem solving task is complicated
by several secondary goals: 1) Match colors within a room, 2) Buy as
much furniture as you can, 3) Spend all your money.  A point system is
used to motivate participants to try to achieve as many goals as possible.
Each subject has a budget and inventory of furniture that
lists the quantities, colors, and prices for each available item.  By
sharing this initially private information, the participants can combine
budgets and select furniture from either's inventory. The problem is
collaborative in that all decisions have to be consensual; funds are
shared and purchasing decisions are joint.

In this context, we characterize an agreement as accepting a partner's
suggestion to include a specific furniture item in the
solution.  In this paper we will focus on the issue of recognizing that a
suggestion has been made (i.e. a proposal).  The problem 
is not easy, since, as speech act theory points out
\cite{austin:1962b,searle75-short}, surface form is not a clear
indicator of speaker intentions.  Consider
excerpt~\refeg{proposal1}:\footnote{We broke the dialogues into
utterances, partly following the algorithm in \cite{pass-manual}.}
\begin{eg}{proposal1}
\item \begin{verbatim}
A: [35]: i have a blue sofa for 300. 
   [36]: it's my cheapest one. 

B: [37]: I have 1 sofa for 350 
   [38]: that is yellow 
   [39]: which is my cheapest,
   [40]: yours sounds good.   
\end{verbatim}
\end{eg}

\noindent
[35] is the first mention of a sofa in the conversation and thus
cannot count as a proposal to include it in the solution.  The sofa A
offers for consideration, is effectively proposed only after the
exchange of information in [37]---[39].
 
However, if the dialogue had
proceeded as below, [35'] would count as a proposal:
\begin{eg}{proposal2}
\item \begin{verbatim}
B: [32']: I have 1 sofa for 350 
   [33']: that is yellow 
   [34']: which is my cheapest.

A: [35']: i have a blue sofa for 300. 
\end{verbatim}
\end{eg}

\noindent
Since context changes the interpretation of [35], 
our goal is to adequately characterize the context.
For this, we look for guidance
from corpus and domain features.
Our working hypothesis is that for both participants context is partly
determined by the domain reasoning situation. Specifically, if the
suitable courses of action are highly limited, this will make an
utterance more likely to be treated as a proposal; this correlation is
supported by our corpus analysis, as we will discuss in
Section~\ref{real-predict}.




\section{Coding Scheme}

\label{scheme}

We will present our coding scheme by first describing the core DRI
scheme, followed by the adaptations for our corpus and research issues.
For details about our scheme, see \cite{coconut-manual}; for details
about features we added to DRI, but that are not relevant for
this paper, see \cite{coconut-ijhcs98}.

\subsection{The DRI Coding Scheme}


The aspects of the core DRI scheme that apply to our corpus are a
subset of the dimensions under \m{Forward-} and \m{Backward-Looking
Functions}.

\subsubsection{Forward-Looking Functions}

This dimension characterizes the potential effect that an utterance
$U_i$ has on the subsequent dialogue, and roughly corresponds to the
classical notion of an \m{illocutionary act}
\cite{austin:1962b,searle75-short}. As each $U_i$ may simultaneously
achieve multiple effects, it can be coded for three different aspects:
\m{Statement, Influence-on-Hearer, Influence-on-Speaker}.

\noindent
{\bf Statement.} The primary purpose of \m{Statements} is to make
claims about the world.  \m{Statements} are subcategorized as an
\m{Assert} when Speaker S is trying to change Hearer H's beliefs, and
as a \m{Reassert} if the claim has already been made in the dialogue.

\begin{figure}
\psfig{figure={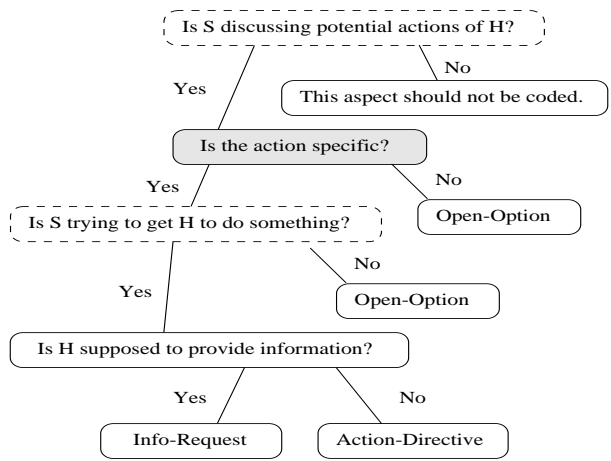},width=8cm,height=6cm,clip=-3cm}
\caption{Decision Tree for Influence-on-Hearer}
\label{fig:future}
\end{figure}

\noindent
{\bf Influence-on-Hearer (I-on-H).}  A U$_i$ tagged
with this dimension influences H's future action.  DRI
distinguishes between S merely laying out options for H's future action
(\m{Open-Option}), and S trying to get H to perform a certain action
(see Figure~\ref{fig:future}). 
\m{Info-Request} includes all actions that
request information, in both explicit 
and implicit forms. 
All other actions\footnote{Although this may cause future problems 
\cite{tuomela:1995}, DRI considers \m{joint} actions as
decomposable into independent \m{Influence-on-Speaker / Hearer}
dimensions.} are \m{Action-Directives}.

\noindent
{\bf Influence-on-Speaker (I-on-S).}
A $U_i$ tagged with this dimension potentially commits S (in
varying degrees of strength) to some future course of action.  The
only distinction is whether the commitment is conditional on H's
agreement (\m{Offer}) or not (\m{Commit}). With an \m{Offer}, S
indicates willingness to commit to an action if H accepts it. 
\m{Commits} include  promises and  other weaker forms.

\subsubsection{Backward Functions} 

This dimension indicates whether
$U_i$ is unsolicited, or
responds to a previous $U_j$ or segment.\footnote{Space constraints
prevent discussion of segments.}  
The tags of interest for our corpus are:\\
$\bullet$ {\bf Answer}: $U_i$ answers a question. \\
$\bullet$ {\bf Agreement}:
\benum
\item $U_i$ \m{Accept/Reject}s if it indicates S's
attitude towards a belief or proposal embodied in its antecedent.
\item  $U_i$ \m{Hold}s if it
leaves the decision about the proposal embodied in its antecedent open
pending further discussion.
\eenum


\subsection{Refinements to Core Features}

The core DRI manual often does not operationalize the tests associated
with the different dimensions, such as the two dashed nodes in
Figure~\ref{fig:future} (the shaded node is an addition that we
discuss below).  This resulted in strong disagreements regarding
Forward Functions (but not Backward Functions) during our initial trials
involving three coders.

\paragraph{Statement.} In the
current DRI manual, the test for \m{Statement} is whether
$U_i$ can be followed by ``That's not
true.''. For our corpus, only syntactic
imperatives or interrogatives were consistently filtered out by this
purely semantic test. Thus, we refined it by appealing to syntax,
semantics, and domain knowledge: $U_i$ is a \m{Statement} if it is
declarative and it is
1) past; or 2) non past, and contains a stative verb; or
3) non past, and contains a non-stative verb in which the 
implied action:\\
$\bullet$ does \m{not} require agreement in the domain;\\
$\bullet$ or is supplying agreement.\\  
For example, \m{We could
start in the living room} is  not tagged as a statement if meant as a
suggestion, i.e. if it requires agreement.

\paragraph{I-on-H and I-on-S.}
These two dimensions depend on the {\em potential action} underlying $U_i$
(see the root node in Figure~\ref{fig:future} for I-on-H). The initial
disagreements with respect to these functions were due to the coders not being
able to consistently identify such actions; thus, we provide a
definition for actions in our domain,\footnote{Our definition of
actions does not apply to \m{Info-Requests}, as the latter are easy to
recognize.} and heuristics that correlate
types of actions with I-on-H/I-on-S.

We have two types of potential actions: \m{put
furniture item X in room Y} and \m{remove furniture item X from room
Y}. We subcategorize them as 
\m{specific} and \m{general}.
A \m{specific}  action  has all necessary
parameters specified ({\em type, price} and \m{color} of 
item, and {\em room}).
\m{General} actions arise 
because all
necessary parameters are not set, as in \m{I have a blue sofa} uttered in
a null context.

\noindent
{\bf Heuristic for I-on-H} (the shaded node in
Figure~\ref{fig:future}). If H's potential action described by 
$U_i$ is \m{specific}, $U_i$ is tagged as 
\m{Action-Directive},  otherwise as \m{Open-Option}.\\
{\bf Heuristic for I-on-S.} Only a $U_i$ that describes
S's \m{specific} actions is tagged with  an {\em I-on-S} tag.

Finally, it is hard to offer comprehensive guidance for the test \m{is S trying
to get H to do something?} in Figure~\ref{fig:future}, but some
special cases can be isolated.  For instance, when S refers to one
action that the participants could undertake, but in the same turn
makes it clear the action is not to be performed, then S is not trying
to get H to do something.  This happens in excerpt~\refeg{proposal1}
in Section~\ref{proposal}. A specific action (\m{get B's \$350
yellow sofa}) underlies [38], which qualifies  as an
\m{Action-Directive} just like [35].  However, because of [40], it is
clear that B is
not trying to get A to use B's sofa. Thus, [38] is 
tagged as an \m{Open-Option}.






\subsection{Coding for problem solving features}

\label{ps-model}

In order to investigate our working hypothesis about the relationship
between context and limits on the courses of action, we coded each
utterance for features of the problem space.  Since we view the
problem space as a set of constraint equations, we decided to code for
the variables in these equations and the number of
possible solutions given all the possible assignments of values to
these variables.

The variables of interest for our corpus are the objects of type \m{t}
in the goal to put an object in a room (e.g. {\em var$_{sofa}$}, {\em
var$_{table}$} or {\em var$_{chairs}$}).  For a solution to exist to
the set of constraint equations, each {\em{var$_i$}} in the set of
equations must have a solution.  For example, if 5 instances of sofas
are known for {\em var$_{sofa}$}, but every assignment of a value to
{\em var$_{sofa}$} violates the budget constraint, then {\em
var$_{sofa}$} and the constraint equations are unsolvable.

We characterize the solution size for the problem as determinate if
there is one or more solutions and indeterminate otherwise.  It is
important to note that the set of possible values for each
{\em{var$_i$}} is not known at the outset since this information must
be exchanged during the interaction.  If S supplies appropriate values
for \m{var$_i$} but does not know what H has available for it then we
say that no solution is possible at this time.  It is also important
to point out that during a dialogue, the solution size for a set of
constraint equations may revert from determinate to indeterminate
(e.g. when S asks what else H has available for a \m{var$_i$}).

\section{Analysis of the Coding Results}

\label{corpus-an}

Two coders each coded 482 utterances with the adapted DRI features
(44\% of our corpus).  Table~\ref{kappa-functions} reports values for
the Kappa (K) coefficient of agreement \cite{Carletta96} for Forward
and Backward Functions.\footnote{For problem solving features, K for two
doubly coded dialogues was $>$ .8.  Since reliability was good and
time was short, we used one coder for the remaining dialogues.}



\begin{table}[t]
\centering
\begin{tabular}{||r|r|r||r|r||}\hline \hline
Stat.  &  I-on-H &  I-on-S  & Answer &  Agr. \\ \hline  \hline 
.83  & .72  & .72  & .79  & .54 
\\ \hline \hline
\end{tabular}
\caption{Kappas for Forward and Backward Functions}
\label{kappa-functions}
\end{table}


The columns in the tables read as follows: if utterance $U_i$ has
tag \m{X}, do coders agree on the subtag? For
example, the possible set of values for \m{I-on-H} are:
NIL ($U_i$ is not tagged with this dimension), Action-Directive,
Open-Option, and Info-Request.  
The last two columns probe the subtypes of Backward Functions: was
$U_i$ tagged as an answer to the same antecedent?  was $U_i$ tagged as
\m{accept}ing, \m{reject}ing, or \m{hold}ing the same
antecedent?\footnote{In general, we consider 2 non-identical
antecedents as equivalent if one is a subset of the other, e.g. if one
is an utterance U$_j$ and the other a segment containing U$_j$.}



K factors out chance agreement between coders; K=0 means agreement is
not different from chance, and K=1 means perfect agreement. To assess
the import of the values $0<K<1$ beyond K's statistical significance (all of
our K values are significant at p=0.000005),
the discourse processing community uses Krippendorf's scale
\shortcite{krippen80}\footnote{More forgiving scales exist but have
not yet been discussed by the discourse processing community, e.g.
the one
in \cite{rietveld93}.}, which
discounts any variable with K $<$ .67, and allows tentative
conclusions when .67~$<$~K~$<$~.8 K, and definite conclusions when
K$\geq$.8.
Using this scale, Table~\ref{kappa-functions} suggests that Forward
Functions and Answer can be recognized far more reliably than
\m{Agreement}.  

To assess the DRI effort, clearly more experiments are
needed. However, we believe our results show that the goal of an
adaptable core coding scheme is reasonable.  We think we achieved good
results on Forward Functions because, as the DRI enterprise intended,
we adapted the high level definitions to our domain. However, we have
not yet done so for Agreement since our initial trial codings did
not reveal strong disagreements; now given our K results, refinement is
clearly needed. Another possible contributing factor for the low K on
Agreement is that these tags are much rarer than the Forward Function tags.
The highest possible value for K may be smaller for low
frequency tags \cite{grove81}.

Our assessment is supported by comparing our results to those of Core
and Allen \shortcite{core-fall97} who used the unadapted DRI manual
--- see Table~\ref{our-roch}.  Overall, our Forward Function results
are better than theirs (the non significant K for I-on-S in
Table~\ref{our-roch} reveals problems with coding for that tag), while
the Backward Function results are compatible.  Finally, 
our assessment may only hold for task-oriented 
collaborative dialogues.  One research group tried to use the DRI core
scheme on free-flow conversations, and had to radically modify it in
order to achieve reliable coding \cite{stolcke-etal}.

\begin{table}[t]
\begin{minipage}{3.125in}
\centering
\begin{tabular}{||r|r|r||r|r||}\hline \hline
Stat.  &  I-on-H & I-on-S & Answer & Agr. \\ \hline \hline
.68 & . 71 & N/S\footnote{N/S means not significant} & .81 & .43 \\ 
\hline \hline
\end{tabular}
\end{minipage}
\caption{Kappas from (Core and Allen 97)}
\label{our-roch}
\end{table}

\section{Tracking Propose and Commit}

\label{real-predict}

It appears we have reached an impasse;
if human coders cannot reliably recognize when two participants achieve
agreement, the prospect of automating this process is grim.  Note that
this calls into question analyses of agreements based on a single
coder's tagging effort, e.g. \cite{walker-ls}.  We think we can
overcome this impasse by exploiting the reliability of Forward
Functions. Intuitively, a  $U_i$ tagged as \m{Action-Directive + Offer}
should correlate with a proposal --- given that all actions in our
domain are joint, an \m{Action-Directive} tag always co-occurs with
either \m{Offer (AD+O)} or \m{Commit (AD+C)}. Further, analyzing the
antecedents of {\em Commit}s should shed light on what was treated as
a proposal in the dialogue. Clearly, we cannot just analyze the
antecedents of \m{Commit} to characterize proposals, as a proposal may
be discarded for an alternative.





To complete our intuitive characterization of a proposal, we will assume
that for a $U_i$ to count as a well-formed proposal (WFP), the context
must be such that enough information has already been exchanged for a
decision to be made.  The feature \m{solution size} represents such a
context.  Thus our first testable characterization of a WFP is: \benum
\item[1.1] $U_i$ counts as a WFP if it is 
tagged as \m{Action-Directive $+$ Offer}  and 
if the associated solution size  is determinate.
\eenum

\begin{table}[t]
\centering
\begin{tabular}{||c||c|c|c||} \hline \hline
& Det & Indet & Unknown \\ \hline
AD$+$O  &  25 & 7 & 0 \\ 
Open-Option & 2 & 2 & 0\\
AD$+$C   & 10 & 2 & 0\\
Other & 4 & 2 & 4 \\ \hline \hline 
\end{tabular}
\caption{Antecedents of Commit}
\label{commit-antec}
\end{table}



To gain some evidence in support of 1.1, we checked whether the
hypothesized WFPs appear as antecedents of
\m{Commits}.\footnote{Antecedents of \m{Commits} are not tagged. We
reconstructed them from either \m{variable} tags or when U$_i$ has
both Commit and Accept tags, the antecedent of the Accept.}  Of the 32
\m{AD+O}s in Table~\ref{commit-antec}, 25 have determinate solution
size; thus, WFPs are the largest class among the antecedents of
\m{Commit}, even if they only account for 43\% of such
antecedents. Another indirect source of evidence for hypothesis 1.1
arises by exploring the following questions: are there any WFPs that
are not committed to? if yes, how are they dealt with in the dialogue?
If hypothesis 1.1 is correct, then we expect that each such
$U_i$ should be responded to in some
fashion.  In a collaborative setting such as ours, a partner cannot
just ignore a WFP as if it had not occurred.
We found that there are 15 \m{AD$+$O}s with determinate solution size
in our data that are not committed to.  On closer inspection, it turns
out that 9 out of these 15 are actually indirectly committed to. Of
the remaining 6, four are responded to with a counterproposal (another
\m{AD$+$O} with determinate solution size).  Thus only two are not
responded to in any fashion.  Given that these 2 occur in a dialogue
where the participants have a distinctively non-collaborative style, it
appears hypothesis 1.1 is supported.

Going back to the antecedents of \m{Commit} (Table~\ref{commit-antec}),
let's now consider the 7 indeterminate \m{AD+O}s.  They can be
considered as \m{tentative} proposals that need to be
negotiated.\footnote{Because of our heuristics of tagging specific
actions as \m{ActionDirectives}, these utterances are not
\m{Open-Options}.}  To further refine our characterization of
proposals, we explore the hypothesis: \benum
\item[1.2] When the antecedent of a \m{Commit} is 
an \m{AD$+$O} and indeterminate, the intervening
dialogue renders the solution size determinate.
\eenum

In 6 out of the 7 indeterminate antecedent \m{AD$+$O}s, our hypothesis is
verified (see  excerpt~\refeg{proposal1}, where [35] is an \m{AD$+$O}
with indeterminate solution size, and the 
antecedent to the \m{Commit} in [40]).

As for the other antecedents of \m{Commit} in Table~\ref{commit-antec}, it is not surprising that
only 4 \m{Open-Options} occur given the
circumstances in which this tag is used (see Figure~\ref{fig:future}).
These \m{Open-Options} appear to function as {\em tentative} proposals
like indeterminate \m{AD$+$O}s, as the dialogue between the
\m{Open-Option} and the \m{Commit} develops according to hypothesis
1.2.
We were instead surprised that \m{AD$+$C}s are a very common category
among the antecedents of \m{Commit} (20\%); the second commit
appears to simply reconfirm the commitment expressed by the first
\cite{walker-phd,walker-ls}, and does not appear to count as a
proposal. Finally, the \m{Other} column is a collection of miscellaneous
antecedents, such as \m{Info-Requests} and cases where the antecedent
is unclear, that need further analysis. For further details, see
\cite{coconut-ijhcs98}.

\section{Future Work}

\label{conclude}


Future work includes, first, further exploring the factors and
hypotheses discussed in Section~\ref{real-predict}. We characterized
WFPs as \m{AD+O}s with determinate solution size: a study of the
features of the dialogue preceding the WFP will highlight how
different options are introduced and negotiated. Second, whereas our
coders were able to reliably identify Forward Functions, we do not
expect computers to be able to do so as reliably, mainly because
humans are able to take into account the full previous context. Thus,
we are interested in finding correlations between Forward Functions
and ``simpler'' tags.


\subsection*{Acknowledgements}

\small This material is based on work supported by the National
Science Foundation under Grant No.~IRI--9314961.  We wish to
thank Liina Pyllk\"{a}nen for her contributions to the coding
effort, and past and present project members Megan Moser and Jerry
Hobbs.

\end{document}